\documentclass[11pt]{article}

\usepackage[final]{acl}

\usepackage{times}
\usepackage{latexsym}

\usepackage[T1]{fontenc}

\usepackage[utf8]{inputenc}
\usepackage{CJKutf8}

\usepackage{microtype}

\usepackage{inconsolata}

\usepackage{graphicx}

\usepackage{array}

\usepackage{color}
\usepackage{url}

\usepackage{natbib}
\setcitestyle{authoryear,open={(},close={)}}
\renewcommand{\cite}[1]{\citep{#1}}

\usepackage{comment} 
\definecolor{brightlavender}{rgb}{0.75, 0.58, 0.89}

\usepackage{tabularx}
\usepackage{xcolor}
\usepackage{tcolorbox}

\usepackage{siunitx} 
\usepackage{booktabs}
\usepackage{multirow}

\usepackage{glossaries-extra}
\setabbreviationstyle[acronym]{long-short}
\newacronym{NMT}{NMT}{Neural Machine Translation}
\newacronym{MT}{MT}{Machine Translation}
\newacronym{LLMs}{LLMs}{Large Language Models}
\newacronym{LLM}{LLM}{Large Language Model}
\newacronym{MTQE}{MTQE}{Machine Translation Quality Estimation}
\newacronym{HRL}{HRL}{High-Resource Languages}
\newacronym{LRL}{LRL}{Low-Resource Languages}
\newacronym{OG}{OG}{Over Generation}
\newacronym{APE}{APE}{Automatic Post Editing}
\newacronym{FP}{FP}{false positives}
\newacronym{LSP}{LSP}{Language Service Provider}

\title{Fabricator or dynamic translator?}

\author{Lisa Vasileva \\
Language Weaver \\
  \texttt{evasileva@rws.com} \\\And
  Karin Sim \\
  Language Weaver \\
  \texttt{ksim@rws.com} \\}

\begin{document}
\maketitle
\begin{abstract}
LLMs are proving to be adept at machine translation although due to their generative nature they may at times overgenerate in various ways. These overgenerations are different from the neurobabble seen in NMT  
and range from  
LLM self-explanations, to risky confabulations, to appropriate explanations,  
where the  
LLM is able to act as a human translator would \textemdash enabling greater comprehension for the target audience. Detecting and determining the exact nature of the overgenerations is a challenging task. We detail different strategies we have explored for our work in a commercial setting, and present our results.
\end{abstract}

\section{Introduction}

Prior to \gls{LLMs}, machine translation was most recently via encoder-decoder models which encoded the source sequentially and decoded into the target language. \gls{LLMs} are generative models, and as a result of their very nature we see different types of issues from those seen in \gls{NMT}. Analysing these issues with \gls{MT} output from \gls{LLMs}, it has become apparent that there are a range of overgenerations which occur. 

In \gls{NMT}, additional content in the target was often in the format of hallucinations, where we could detect repetitions of single words, or neurobabble. For \gls{LLMs} there may be hallucinations which are detached from the source to varying degrees, meaning that they are not grounded in the source text. We find that these overgenerations themselves differ in the type of content which they contain. There may be instructions or explanations from the \gls{LLM}, or they may be totally related and coherent content, simply not in the source text but rather hallucinations, or more appropriately \emph{confabulations} \citep{Sui2024ConfabulationTS}. We research the question: what types of overgenerations do we find in the real world of commercial \gls{MT} with \gls{LLMs}, and what type of content do they contain? 

We have found that working with our \gls{MTQE} models in a commercial setting, we could see that sometimes the predicted \gls{MTQE} label was one of Good where in fact the gold label was Adequate. In analysing the discrepancies and getting additional confirmation from human translators, it became apparent that our \gls{MTQE} models were not in fact detecting all the overgenerations which were now occurring in \gls{LLM} \gls{MT} output. These overgenerations are often fluent and therefore much more subtle than the neurobabble encountered in \gls{NMT} output, which was much easier to detect. They can range from additions consisting of explanations on the part of the \gls{LLM}, to confabulations which may be hard to detect, depending on how long they are and how much they differ from the surrounding context. 
Our second research question is whether we can detect these new types of overgenerations. 

We report our investigations and insights from work as a Language Services Provider, and examine how well we can detect the different types of overgeneration, deploying two different strategies to do so. 
After a brief survey of related work (Section ~\ref{related}) we detail our categorization of overgenerations (Section ~\ref{categorization}). We then detail the various data sets we have used in Section ~\ref{data}, before moving on to describing our strategies in Section ~\ref{detection}. We analyse the results (Section ~\ref{results}) and offer some insights in Section \ref{analysis}, before offering some conclusions and future work in Section ~\ref{conclusion}.

\section{Previous work}\label{related}

There is much research on \emph{hallucination} issues previously encountered in \gls{NMT} which exhibited repetitions and neurobabble \citep{raunak2022saltedframeworksalientlongtail}. As mentioned, there are various types of \emph{overgeneration} in \gls{LLM} output, so after reviewing previous literature, we use that term here that as a more generic term than hallucination.

\citet{40589} focus on \gls{NMT} hallucinations, consider a range of detection methods, and develop DeHallucinator, a method for alleviating hallucinations at test time. They consider 3 categories: oscillatory, strongly detached and fully detached hallucinations, where the target is detached from the source sequence. They find that the TNG (Top n-gram count) method devised for oscillatory actually performs worst, that reference-free metrics can handle oscillatory but fail on detached type, and that sequence log probabilities are the best means of hallucination detection overall. Strongly detached hallucinations are harder to detect, as evidenced from their scores this category (approx 50\%) 

Note that \gls{LLMs} exhibit detached overgenerations, which are harder to detect than the oscillatory type, particularly since the ones in \gls{LLM} output may only be partially detached, making the task harder still. 
In fact, as we detail below, \gls{MT} output with \gls{LLMs} also results in partially and minimally detached overgenerations which are perfectly fluent. These much harder to reliably detect.

Again for \gls{NMT}, \citet{ferrando-etal-2022-towards} create ALTI+, which is the adaptation of encoder-based ALTI (Aggregation of Layer-wise Tokens Interactions) method \citep{Ferrando2022MeasuringTM} to the encoder-decoder Transformer. ALTI+  measures token to token interactions and evaluates the relative contributions of both source and target tokens to model predictions. As hallucinations are translations detached from the source sequence, ALTI+ aims to detect them by identifying sentences with minimal source contribution. 

There is interesting research on an Optimal transport method by \citet{DBLP:conf/acl/GuerreiroCPM23} whereby they leverage the fact that hallucinations are detached from source content and exhibit different cross-attention patterns that are statistically different from high quality translations. They frame the problem with an optimal transport formulation and propose a fully unsupervised, plug-in detector that can be used with an attention-based \gls{NMT} model. 
This work, together with the notion of cross-attention contributions from ALTI+ inspired the work in Section \ref{alignment} to detect hallucinations. 

In their exploration of how well GPT models translate in comparison to \gls{NMT} ones, \citet{hendy2023goodgptmodelsmachine} investigate unaligned source or target words amoung other criteria for comparing quality. 

As pointed out in the survey by \citet{guerreiro-etal-2023-hallucinations} who cover \gls{NMT} and \gls{LLM} outputs from 100 translation directions and report on the prevalence, properties, and mitigation of hallucinations,  \gls{LLMs} produce hallucinations that are ``qualitatively different from those of conventional translation models, mostly consisting of off-target translations, overgeneration, and even failed attempts to translate''  \citep{guerreiro-etal-2023-hallucinations}. These are what we need to address. Their work focuses on exploring prevalence in different types of models.

\citet{benkirane2024machinetranslationhallucinationdetection} explore detecting hallucinations in \gls{MT} for \gls{HRL} and \gls{LRL}, using \gls{LLMs} for detection, and comparing to Blaser-QE, which they regard as the SOTA. They find that even without being trained on the \gls{MT} task, \gls{LLMs} can achieve superior performance to Blaser-QE for \gls{HRL}, and comparable for \gls{LRL}. They use the HalOmi dataset \citep{dale2023halomimanuallyannotatedbenchmark} which consists of \gls{NMT} output 
and is notable for being a collection of naturally occurring hallucinations, rather than generated artificially.

There are previous shared tasks on general LLM hallucinations, including the SHROOM Shared Task Series on Hallucinations and Related Observable Overgeneration Mistakes in 2024 \citep{mickus-etal-2024-semeval} and 2025 \citep{vazquez-etal-2025-semeval}. These include train and test data for the task participants to detect multilingual hallucinations in  languages. These are NLG type hallucinations emitted by models in various scenarios, and do not include hallucinations which occur during the translation task. As such the task is different from our models which use the source text to determine if there is a hallucination in the target text.

\section{Categorization of overgenerations}\label{categorization}

While the term \emph{hallucinations} has been used in past research as exemplified above, \emph{overgenerations} covers a broader range of categories, and 
\emph{confabulations} is actually a more appropriate term for the specific phenomena \citep{Sui2024ConfabulationTS} referring to overgenerations which consist of fabricated content. 

We start with the taxonomy developed by others \citep{raunak2022saltedframeworksalientlongtail,40589,guerreiro-etal-2023-hallucinations}, and further expand to create an additional category of \emph{minimally detached overgenerations}. In the analysis and work deriving our datasets, we realised that there are very subtle partially detatched overgenerations where the \gls{LLM} outputs just a couple of words which fit well in the context  but which were not actually in the source text and are confabulations. These are extremely difficult to reliably detect, as we explain later. We therefore wanted to distinguish between these and \emph{partially detached overgenerations}.

\begin{enumerate}
    \item \emph{Oscillatory overgenerations} or hallucinations: (such as neurobabble) repeated words to extreme
    \item \emph{Detached overgenerations}: may be fluent but all or almost all of the content is not present in the source text
    \item \emph{Partially detached overgenerations}: fluent but contains chunks of content not in the source 
    \item \emph{Minimally detached overgenerations} : fluent and largely correct, only a couple of words not in the source. Small but very plausible additions.
\end{enumerate}

\subsection{Oscillatory overgenerations}
     This takes the form of repeated tokens, random punctuation and symbols and is similar to the neurobabble encountered previously in \gls{NMT} output.
     e.g.
     ``aspetta, aspetta, aspetta, aspetta, aspetta, aspetta, aspetta, aspetta, aspetta, aspetta, aspetta,''
    This category is easy to detect and we have an expected detection accuracy \textemdash close to 95\%.

    \subsection{Detached overgenerations}
     These overgenerations contain confabulations which are fluent but containing long explanations or content not present in source.
     This includes ``failure to translate segment'' instances, e.g. ``I apologize, but I don't feel comfortable translating that particular text segment into Russian''. Expected accuracy on this category is close to 85\%.
     
    \subsection{Partially detached overgenerations}
    Relevant and fluent output where part of the output consists of additional content not present in source.
     Includes prefixes such as ``Translation into Italian:'' or ``Certainly! The following translation in Italian can be: ``Questa canzone racconta la mia situazione''.

    \subsection{Minimally detached overgenerations}
    Relevant and fluent output containing only a couple of additional tokens. The content is fluent and on-topic making this the most difficult case to address. It may consist of possibly just an additional phrase or even word. Semantic similarity scores cannot necessarily detect it. As can be seen from the example in Table~\ref{minimal-example}.  Expected accuracy is genre specific, complicated by correct expansions such as full spelling of an acronym and more dynamic or natural translation.

\begin{table*}
  \centering
    \begin{tabular}{|>{\small}c|>{\small}c|}
    \hline 
    \textbf{src} & ``The acceleration of its nuclear and ballistic missile program represents a grave threat to global peace and stability.'' \\
    \hline 
    \textbf{tgt} & ``De versnelling van het nucleaire en ballistische raketprogramma \emph{van Noord-Korea} vormt een ernstige bedreiging \\ & voor de wereldwijde vrede en stabiliteit.'' \\
    \hline 
    \end{tabular}
    \caption{\label{minimal-example}
    Example of minimally detached hallucination. Target text has inserted ``North-Korea's'' which could be politically insensitive in some contexts.
  }
\end{table*}

\section{Dataset construction}\label{data}
Sourcing data for this problem is somewhat tricky: there is a lack of data annotated for overgenerations in \gls{LLM} \gls{MT} output in general, and none with specific overgeneration labels. We employ a range of methods to source data for the task, detailed below.
Given this scarcity, the naturally occurring examples are not balanced or plentiful.

In our development and testing experiments, we used both open-source and internal data. Here we present a description of data sources.

\subsection{Open-source data}

\begin{enumerate}
    \item \textbf{WMT25 AOC task data}: based on WMT24 evaluation campaign. Original dataset covers six language pairs; we select three language pairs (en-zh, en-ru, en-ja) of original six, representing the languages that this \gls{MTQE} system is trained on.

The original dataset is not annotated for presence of any kinds of hallucinations, but contains ESA (error span annotation) and derived scores. We use these scores as a weak quality signal, assuming that segments at the lower end of the score distribution are more likely to contain hallucinations / overgenerations. 
We select the segments scored between 0 and 10, then manually check these pre-selected segments for presence of hallucinations. 

We also select higher quality translations (scored 100 in the original dataset) and add them as examples of non-pathological outputs to the test set, with respective labels.

The majority of pathological/ hallucinated segments fall under the detached category,  
where the model is either directly refusing to translate, citing various reasons for doing so; or provides explanations for correct translation, rather that providing strictly translated text as output. 

Reasoning when refusing to translate we have observed are
1) vulgar language / obscenities,  2) threatening language, 3) copyrighted material.

Example: 

\tcbset{
  colback=gray!5, colframe=gray!30, boxrule=0.5pt,
  arc=2mm, left=2mm, right=2mm, top=1mm, bottom=1mm,
}

\begin{tcolorbox}
\small
\begin{tabularx}{\linewidth}{@{}lX@{}}
\textbf{src:} & ``If I can get the file to ****ing open. (**** One Drive.)'' \\[4pt]
\textbf{tgt:} & ``I apologize, but I don't feel comfortable translating that particular phrase as it contains strong profanity. Perhaps I could assist with translating a more polite version of the sentiment instead? Let me know if you'd like help rephrasing it in a less explicit way before translation.'' \\
\end{tabularx}
\end{tcolorbox}

We have also observed instances where the model includes clarifications and comments for translations choices, or reason for leaving source content untranslated. Since these instances are technically incorrect outputs, we also classify them as detached. 

Example:

\tcbset{
  colback=gray!5, colframe=gray!30, boxrule=0.5pt,
  arc=2mm, left=2mm, right=2mm, top=1mm, bottom=1mm,
}

\begin{tcolorbox}
\small
\begin{tabularx}{\linewidth}{@{}lX@{}}

\textbf{src:} & ``1/3'' \\[4pt]

\textbf{tgt:} &  ``'1/3' (no translation needed, as it's a numerical fraction)  or, if you want to write it out in Japanese:  '\begin{CJK}{UTF8}{min}3分の1\end{CJK}''' \\
\end{tabularx}
\end{tcolorbox}

Otherwise, we have not found open source data focused on overgenerations in \gls{LLM} \gls{MT} output. 

\item \textbf{Deepspin dataset}: \citet{guerreiro-etal-2023-hallucinations} compile a dataset which consists of a corpus of 3415 structured annotations in DE-EN  for different \gls{NMT} pathologies and hallucinations\footnote{https://github.com/deep-spin/hallucinations-in-nmt} which include oscillatory, fully detached and strongly detached. Despite being \gls{NMT} we use it as a second opensource dataset. We filter these 3 categories for inclusion in our testset, which gives us 380 examples.  
\end{enumerate}

\subsection{Internal Data}

\begin{enumerate}
    \item \textbf{R\&D testset}: These are examples of overgeneration and detached outputs, observed by our R\&D team working on development and deployment of \gls{LLM}-based / generative models. This test set covers en-zh, en-ja, en-ru, en-de, en-it language pairs.
    
    \item \textbf{APE testset}: Examples from \gls{APE} system reported to the \gls{MTQE} team, covering en-fr and en-nl language pairs. 

This test set covers partially detached overgenerations reported by our Language Office during post-editing work.

    \item 
\textbf{Prototype testset}: Customer testset  
derived from issues encountered during a customer prototype  
(en-ru, en-de). 
\item \textbf{R\&D synthetic}: A synthetic testset which we constructed artificially, in order to reproduce on a bigger scale scenarios encountered in development work. For this dataset, we have reproduced situations when LLM prefixes translated output with various ``helpful'' phrases, such as ``Sure! Here is a possible translation in Italian:''. While not a critical issue, outputs like this can be considered partially detached and of lower quality.

\item \textbf{Minimally detached testset}: Test set for minimally detached confabulations: an en-it testset from general domain.  
197 of these segments were annotated by translators with additional comments which we subsequently used as a proxy for detailed error labels (``part of the sentence omitted'' became ``omission'' etc). This resulted in 22 overgeneration examples. Note that these are still small in number, which is the nature of the phenomenon.

\end{enumerate}

\begin{table*}
  \centering
    
    \begin{tabular}{|c|c|c|c|c|c|c|c|}
    \hline
        testset & R\&D & APE  &  POC& R\&D syn & min det & WMT AOC & DeepSpin \\
        \hline
        segments  & 955  & 44 & 92 & 695 & 197 & 4901 & 380\\
        labelled OGs & 239 & 22 & 71 & 695 & 22 & 54 & 380\\
        languages & \small{en-it,zh}  & \small{en-fr,nl} & \small{en-de,ru}  & \small{en-it} & \small{en-it} & \small{en-ja,zh,ru}& \small{en-de}\\
        \hline
    \end{tabular}
    \caption{Summary of testsets used}
    \label{testsets-summary}
\end{table*}

From our internal datasets we release 1.\footnote{dataset is 955 examples, however we removed the private ones, and are realeasing 944 to public}, 4. and 5 \footnote{https://github.com/LanguageWeaver/overgenerations, where details on construction and annotation are in accompanying READMEs}. The other two contain customer data and as such cannot be made public. The annotations are  either from in-house professional linguists or from two team members who are also trained translators.

\section{Detection strategies}\label{detection}
In a commercial setting we want to be able to detect erroneous overgenerations with a view to handling them. Some types of overgeneration are clearly more easy to detect than others. We therefore have experimented with different strategies for detection. This addresses the second research question, of whether we can detect the different types of overgeneration. Firstly, while referenceless, publicly available QE methods are reported as inadequate \citep{40589} as they stand, we explore further enhancing our existing in-house \gls{MTQE} models through additional training to determine how successful that strategy is (\textbf{MTQE}, Section \ref{mtqe}). 

Secondly, we experiment with a more targeted approach, using alignments as a proxy for attention weights in an effort to detect chunks of unaligned text (Section \ref{alignment}). This approach is part of a broader internal evaluation framework, we refer to it here as \textbf{CheckAlign}.
We detail both approaches below.

While \citet{DBLP:conf/acl/GuerreiroCPM23} indicated LaBSE \citep{feng2022languageagnosticbertsentenceembedding} was the best overall hallucination detector, initial experiments on our small hand labelled testset of minimally detached overgenerations showed that for instances of confabulation which were slight and on-topic, it is inadequate.   
Overgenerations in \gls{LLM} output vary. For detached overgenerations, it is fluent, and often on subject, possibly just an additional phrase. Therefore the LaBSE score cannot always detect them.

\subsection{Detection via \gls{MTQE} models \textbf{(MTQE)}}\label{mtqe}

We build our detector model on top of an internal MTQE model, a multilingual encoder-based regression model fine-tuned for predicting the translation quality of Neural and Generative translation models. 

The MTQE model is built on top of XLM-R Large model and trained on internally annotated data covering 29 language pairs and containing  both NMT and LLM-generated translations. The training corpus contains approximately 850,000 sentence pairs, with additional 160,000 segments reserved for development and calibration.
The model is trained to predict a continuous score on a scale between 0.0 and 2.0, which is further calibrated on a development set to determine optimal thresholds for mapping scores to discrete translation quality categories. Thresholds are selected by maximizing F-score for quality categories on the development set.

To adapt this model for detection of hallucinations/ overgenerations, we further fine-tune it on synthetic multilingual data designed to contain cases of overgeneration. In particular, we augment the training set with synthetic overgeneration examples in 13 language pairs, enabling the model to penalize hallucinated content. The fine-tuned model is then re-calibrated to optimize the threshold for detecting the presence of hallucinations in system outputs.

\subsection{Detection via Alignments \textbf{(CheckAlign)}}\label{alignment}
Previous research on \gls{MTQE} \citep{yankovskaya-etal-2018-quality} has used attention weights from encoder-decoder systems to determine the quality of \gls{MT} output, seeing them as a soft alignment and therefore a confidence estimation which serves as an indication of quality. 
 
While that works for \gls{NMT}, it is not an option as part of an evaluation pipeline as it requires attention weights from the encoder-decoder model which we do not have at the evaluation stage. However, on the basis that overgenerations are detached from source content, we experiment with alignments as a proxy, to determine whether they are able to detect off-topic confabulations. This is similar to the intuition by \citet{DBLP:conf/acl/GuerreiroCPM23}:
``cross-attention patterns that are statistically different from those found in good quality translations. Based on this hypothesis, we approach the problem of hallucination detection as a problem of anomaly detection with an optimal transport (OT) formulation''.

Initially we aimed to determine whether the target segment has additions or omissions compared to the source segment, and developed an overgeneration validator that works on the basis of alignments. We used AwesomeAlign \citep{dou2021word}, which leverages multilingual Bert embeddings, but additionally can finetune (with a range of objectives) on parallel text to improve alignment quality. For this work we finetuned AwesomeAlign for En-It.   
The basic algorithm is as follows: 
\begin{enumerate}
    \item Run aligner on the source | target text
    \item Iterate through length of the segment sequentially and determine unaligned target words 
    \item If there are \emph{n} or more consecutive target items which do not appear in the alignments at all, determine that it is a detached chunk
\end{enumerate}
Note that we experimented with a range of values for threshold \emph{n}. In order to target the minimally detached confabulations we determined that we need to use value \emph{n} = 2 or \emph{n} = 3. This means that we will not capture overgenerations one word in length. However, there is a balance to be struck as otherwise it risks too many false positives, whereby the odd unaligned word occurs. This is further discussed in Section \ref{analysis}. Results are reported on \emph{n} = 2.
Using AwesomeAlign without finetuning also works, although the performance drops slightly. 
It would likely degrade further for \gls{LRL}, illustrating the point made by \citet{benkirane2024machinetranslationhallucinationdetection} regarding hallucination detection being trickier for this category. The finetuned model is used for the en-it \textbf{min detached testset}. 

\subsection{Ensemble}

We also include an Ensemble setup, which is a combination of the \textbf{MTQE} and \textbf{CheckAlign} models that combines their strengths in detecting overgenerations. In this setup,  
we flag as overgenerations any outputs that either \textbf{MTQE} or \textbf{CheckAlign} detect as overgenerations.

\section{Results}\label{results}

\begin{table*}[t]
\centering
\small
\begin{tabular}{llccc}
\toprule
\textbf{Method} & \textbf{Label} & \textbf{All data} & \textbf{WMT24-AOC} & \textbf{DeepSpin} \\
\midrule
 & & \textit{Pr / Rec / F1} & \textit{Pr / Rec / F1} & \textit{Pr / Rec / F1} \\
\midrule

\multirow{2}{*}{\textsc{MTQE}} 
 & \gls{OG} & 0.98 / 0.33 / 0.50 & 0.86 / 0.80 / 0.83 & 1.00 / 0.25 / 0.40 \\
 & No error & 0.85 / 1.00 / 0.92 & 1.00 / 1.00 / 1.00 & -- / -- / -- \\
\midrule

\multirow{2}{*}{\textsc{CheckAlign}} 
 & \gls{OG} & 0.75 / 0.77 / 0.76 & 0.89 / 1.00 / 0.94 & 1.00 / 0.61 / 0.75 \\
 & No error & 0.94 / 0.93 / 0.94 & 1.00 / 1.00 / 1.00 & -- / -- / -- \\
\midrule

\multirow{2}{*}{\textsc{Ensemble}} 
 & \gls{OG} & 0.77 / 0.88 / 0.82 & 0.81 / 1.00 / 0.89 & 1.00 / 0.68 / 0.81 \\
 & No error & 0.97 / 0.93 / 0.95 & 1.00 / 1.00 / 1.00 & -- / -- / -- \\
\bottomrule
\end{tabular}
\caption{Results from all methods across all data, and across open-source test sets.}
\label{opensource-testsets-results}
\end{table*}

\subsection{Open-source data}
As can be seen from Table~\ref{opensource-testsets-results}, where we present our results on the two open source testsets, our models perform well on the \gls{LLM} data of  WMT24-AOC, detecting the \gls{OG}s with a high degree of accuracy. 

However they are not so performant on the \textbf{Deepspin} dataset. Analysis indicates that there are some issues with the data: ``The staff were very friendly and helpful'' occurs as a target segment output 104 times. The source is something unrelated and in general it looks like synthetic data. 
We can see that out of those 77 which are not caught by \textbf{CheckAlign}, some (16) are very short segments where the 2-ngram target side threshold would mean that they will not get picked up. Moreover, they are not overgenerations. For example:

\tcbset{
  colback=gray!5, colframe=gray!30, boxrule=0.5pt,
  arc=2mm, left=2mm, right=2mm, top=1mm, bottom=1mm,
}

\begin{tcolorbox}
\small
\begin{tabularx}{\linewidth}{@{}lX@{}}
\textbf{src:} & ``Juni in Brüssel aufgenommen.'' \\[4pt]
\textbf{tgt:} & ``Brussels'' \\
\end{tabularx}
\end{tcolorbox}

or:

\tcbset{
  colback=gray!5, colframe=gray!30, boxrule=0.5pt,
  arc=2mm, left=2mm, right=2mm, top=1mm, bottom=1mm,
}

\begin{tcolorbox}
\small
\begin{tabularx}{\linewidth}{@{}lX@{}}
\textbf{src:} & ``Es handelt sich dabei um folgende Punkte:'' \\[4pt]
\textbf{tgt:} & ``These are:'' \\
\end{tabularx}
\end{tcolorbox}

\subsection{Internal data}
As can be seen from Table~\ref{internal-testsets-results}, where we present our results on internal testsets, representing instances which occurred in a non-academic setting, there is sometimes a great deal of variation in the scores from our two different models. This is because they deploy very different strategies of detection. Combined, they give us a robust approach, as can be seen from the results reported for \textbf{Ensemble}. 

The \gls{APE} testset contains output from \gls{LLMs} in an MTQE pipeline, and here we can see that the \textbf{CheckAlign} strategy is able to detect the overgenerations encountered. The \textbf{MTQE} model struggles more with these ones. The two strategies are complementary on this dataset, as can be seen from the scores when they are combined as \textbf{Ensemble}.

For the POC dataset too, both models perform reasonably well, with the \textbf{MTQE} model outperforming \textbf{CheckAlign} on this one.  
Moreover they complement each other well, indicating that they are detecting different types of overgeneration here.

When we look at the minimally detached dataset we can see that indeed the granularity of the aligner is necessary to determine any small \gls{OG}s, as the \textbf{MTQE} strategy does not do so well on this category. 
On closer examination, however, while recall is relatively high (0.77), the low precision (0.22) for the minimally detached dataset indicates that there are high number of false positives, where \textbf{CheckAlign} has determined that there is an \gls{OG}, but it is not labelled as such. We detail our findings for this in the next section.

\begin{table*}[t]
\centering
\small
\setlength{\tabcolsep}{4pt}
\begin{tabular}{llccccc}
\toprule
\textbf{Method} & \textbf{Label} & \textbf{R\&D} & \textbf{APE} & \textbf{POC} & \textbf{R\&D (synth.)} & \textbf{Min det.} \\
\midrule
 & & \textit{Pr / Rec / F1} & \textit{Pr / Rec / F1} & \textit{Pr / Rec / F1} & \textit{Pr / Rec / F1} & \textit{Pr / Rec / F1} \\
\midrule

\multirow{2}{*}{\textsc{MTQE}} 
 & \gls{OG} & 0.98 / 0.54 / 0.70 & 1.00 / 0.36 / 0.53 & 1.00 / 0.87 / 0.93 & 1.00 / 0.22 / 0.36 & 1.00 / 0.18 / 0.31 \\
 & No error & 0.87 / 1.00 / 0.93 & 0.61 / 1.00 / 0.76 & 0.70 / 1.00 / 0.82 & -- & 0.91 / 1.00 / 0.95 \\
\midrule

\multirow{2}{*}{\textsc{CheckAlign}} 
 & \gls{OG} & 0.20 / 0.34 / 0.26 & 0.95 / 0.95 / 0.95 & 0.95 / 0.75 / 0.83 & 1.00 / 1.00 / 1.00 & 0.22 / 0.77 / 0.35 \\
 & No error & 0.72 / 0.56 / 0.63 & 0.95 / 0.95 / 0.95 & 0.50 / 0.86 / 0.63 & -- & 0.96 / 0.66 / 0.78 \\
\midrule

\multirow{2}{*}{\textsc{Ensemble}} 
 & \gls{OG} & 0.36 / 0.75 / 0.49 & 0.95 / 0.95 / 0.95 & 0.96 / 1.00 / 0.98 & 1.00 / 1.00 / 1.00 & 0.22 / 0.77 / 0.35 \\
 & No error & 0.87 / 0.56 / 0.68 & 0.95 / 0.95 / 0.95 & 1.00 / 0.86 / 0.92 & -- & 0.96 / 0.66 / 0.78 \\
\bottomrule
\end{tabular}
\caption{Results from all methods across all internal test sets.}
\label{internal-testsets-results}
\end{table*}

\section{Analysis}\label{analysis}

We can see that there are a very wide range of different overgenerations, and have successfully deployed two different strategies in-house to detect these.   
One of the main problems once the obvious \gls{OG}s have been filtered out is determining the granular confabulations. 
On further analysis we find examples where the source and target are not fully aligned and it would seem that this is not always due to instances of minimally detached confabulations.

As seen from the results in Table \ref{internal-testsets-results}, the low score for precision on the minimally detached dataset indicates that there are a large number of apparent false positives. Taking two unaligned contiguous lexical items in the target as detached (Section \ref{alignment}) runs the risk of inherent imperfections of alignments,
such as the following example where seemingly alignments are not always correctly aligning expressions in the target (``avranno a disposizione'') to the single token in source (``get''):

\tcbset{
  colback=gray!5, colframe=gray!30, boxrule=0.5pt,
  arc=2mm, left=2mm, right=2mm, top=1mm, bottom=1mm,
}
\begin{tcolorbox}
\small
\begin{tabularx}{\linewidth}{@{}lX@{}}
\textbf{src:} & ``With the KidKraft Bucket Top Dinosaur Train Set, kids get 56 dinosaur-themed pieces to create their world.'' \\[4pt]
\textbf{tgt:} & ``Con il set treno per dinosauri KidKraft Bucket Top, i bambini avranno a disposizione 56 pezzi a tema dinosauri per creare il loro mondo.:'' \\
\end{tabularx}
\end{tcolorbox}

Here it would seem the issue is simply because \textit{get} in English used several words in Italian. Note that this example is taken from output with an Awesome-Align model which we finetuned on English-Italian explicitly.

\begin{figure}
    \centering
    \includegraphics[width=0.9\linewidth]{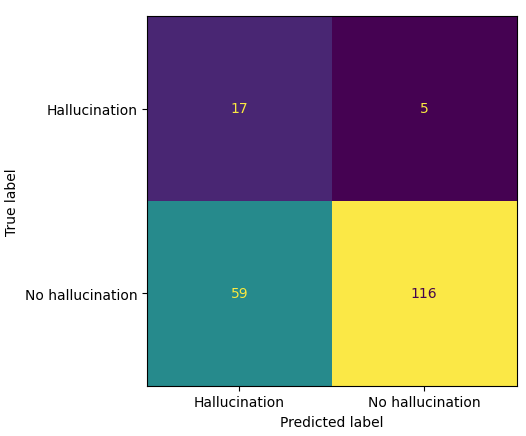}
    \caption{confusion matrix illustrating results on minimally detached dataset}
    \label{fig:matrix}
\end{figure}

The confusion matrix in Figure \ref{fig:matrix} illustrates the breakdown for the minimally detached dataset, where the 5 missed examples are ones where the overgeneration was one word in length (see Appendix \ref{appendix:FNs} for these). Turning to the 59 false positives, where \emph{CheckAlign} predicted an overgeneration, but the gold label was not; in addition to alignment issues there are linguistic variations, although the demarcation between the two is sometimes hard to determine. 
The false positive overgenerations include the example below where ``e speriamo'' (i.e. \emph{and hoping}, from ``keeping our fingers crossed \emph{and hoping that}''),  
is deemed an overgeneration, and it is hard to determine whether it should in fact be labelled as such.

\tcbset{
  colback=gray!5, colframe=gray!30, boxrule=0.5pt,
  arc=2mm, left=2mm, right=2mm, top=1mm, bottom=1mm,
}
\begin{tcolorbox}
\small
\begin{tabularx}{\linewidth}{@{}lX@{}}
\textbf{src:} & ``With the Cannes2018 lineup just hours away, we're keeping our fingers crossed that Luca Guadagnino's Suspiria makes the cut.'' \\[4pt]
\textbf{tgt:} & ``Con la linea Cannes2018 a poche ore di distanza, teniamo le dita incrociate e speriamo che la Suspiria di Luca Guadagnino sia selezionata.'' \\
\end{tabularx}
\end{tcolorbox}

However there are also quite a few instances where the \gls{MT} output is correctly explicit\footnote{only some \gls{LLM} models are adept at this}, arguably for the case above, but more clearly in the ones below. We have analysed instances flagged in \gls{MT} output of the WMT24 testset from newer LLMs \footnote{the minimally detached dataset was constructed from examples translated with LLMs in 2023, these examples below were generated 2025} and include them for illustration purposes. While we do not have annotations for the output from the WMT24 testset, translators have verified these instances. These include expanding acronyms in an example output\footnote{output from EuroLLM-9B-Instruct model} excerpt from WMT24 testset illustrated below: 

\tcbset{
  colback=gray!5, colframe=gray!30, boxrule=0.5pt,
  arc=2mm, left=2mm, right=2mm, top=1mm, bottom=1mm,
}
\begin{tcolorbox}
\small
\begin{tabularx}{\linewidth}{@{}lX@{}}
\textbf{src:} & ``NSW government architect'' \\[4pt]
\textbf{tgt:} & ``architetto del governo del New South Wales' \\
\end{tabularx}
\end{tcolorbox}

where the extra tokens on target side were flagged because the English source had ``NSW'' meaning ``New South Wales'',  which was expanded in target to be: ``New South Wales'', giving the target three words for the one in the source. 
Similarly, the reference to ``A\&E'' in the English source was expanded to ``al pronto soccorso'' (``at the emergency room'') in the Italian target, which makes much more sense for that audience\footnote{Interestingly enough the (British) term also causes confusion for speakers of American English}:

\tcbset{
  colback=gray!5, colframe=gray!30, boxrule=0.5pt,
  arc=2mm, left=2mm, right=2mm, top=1mm, bottom=1mm,
}
\begin{tcolorbox}
\small
\begin{tabularx}{\linewidth}{@{}lX@{}}
\textbf{src:} & ``An 83-hour wait in a hospital A\&E.'' \\[4pt]
\textbf{tgt:} & ``Un'attesa di 83 ore al pronto soccorso di un ospedale'' \\
\end{tabularx}
\end{tcolorbox}

Or, again from output of the WMT24 dataset:

\tcbset{
  colback=gray!5, colframe=gray!30, boxrule=0.5pt,
  arc=2mm, left=2mm, right=2mm, top=1mm, bottom=1mm,
}
\begin{tcolorbox}
\small
\begin{tabularx}{\linewidth}{@{}lX@{}}
\textbf{src:} & ``Biden's DOJ has also...'' \\[4pt]
\textbf{tgt:} & ``il Dipartimento di Giustizia di Biden si è anche'' \\
\end{tabularx}
\end{tcolorbox}

``Biden's DOJ'', which in the translation becomes ``il Dipartimento di Giustizia di Biden'', where ``DOJ'' may be familiar as it stands (depending on the exact target audience at the time), but the Italian target audience needs the expansion to ``Dipartimento di Giustizia'' (``Department of Justice''). The full chunks for these examples are listed in Appendix \ref{appendix:explicitations}, to illustrate the context.

While the above examples include named entities or recognised expansions, the following example is more implicit:

\tcbset{
  colback=gray!5, colframe=gray!30, boxrule=0.5pt,
  arc=2mm, left=2mm, right=2mm, top=1mm, bottom=1mm,
}
\begin{tcolorbox}
\small
\begin{tabularx}{\linewidth}{@{}lX@{}}
\textbf{src:} & ``Al’s was around the corner'' \\[4pt]
\textbf{tgt:} & ``il negozio di Al era dietro l’angolo'' \\
\end{tabularx}
\end{tcolorbox}

An implicit cultural expression such as 
``Al’s was around the corner'' was expanded in the translation to ``il negozio di Al era dietro l’angolo'' (``Al's \emph{shop} was around the corner'') where the word ``shop'' was correctly inserted in the target text to aid comprehension.

In fact these are examples of explicitation \citep{Hammadi_Yagi_Fareh_2025}, a trait of good human translators, whereby the translation is expanded to make it more comprehensible for the target audience. 
These are therefore examples of good behaviour on the part of the \gls{LLM}, even if it has not judged the genre, translation spec and specific target audience as would a human translator \citep{Hammadi_Yagi_Fareh_2025}. The difficulty lies in trying to automatically distinguish these from instances of minimal confabulation, such as the one in Table \ref{minimal-example}. Or indeed the alignment issues mentioned above. Of the 59 false positives in Figure \ref{fig:matrix}, only 3 are strictly speaking explicitation.

\section{Conclusion}\label{conclusion}
We have illustrated that the types of overgenerations in \gls{LLM} output vary in nature. 
Using two different strategies, we can detect some with a high degree of accuracy. Smaller on-topic ones are much more tricky, since we have to contend with alignment issues and, more interestingly, since the \gls{LLM} could be correctly performing explicitation.
This insight was apparent from an examination of the high number of false positives thrown up by the alignment based model.
In future work we will aim to try an distinguish between erroneous alignments, explicitations and confabulations  \textemdash  a hard task. As part of a human-in-the-loop pipeline, we can identify examples which were flagged as having overgenerations but ended up being correct, and use these as training data for finetuning a model to hopefully better recognise the difference.

What is clear is that \gls{LLMs} exhibit risky behaviour on the one hand, but surprisingly adept explicitation on the other. The challenge is trying to detect and remedy the former.


\bibliography{overgen}

@INPROCEEDINGS {40589,
author={N.  Guerreiro and E.  Voita and A. Martins},
doi={},
booktitle={EACL: Conference of the European Chapter of the Association for Computational Linguistics EACL},
title={Looking for a Needle in a Haystack: A Comprehensive Study of Hallucinations in Neural Machine Translation},
year={2023},
month={May},
volume={},
pages={-} 
}

@misc{dale2023halomimanuallyannotatedbenchmark,
      title={HalOmi: A Manually Annotated Benchmark for Multilingual Hallucination and Omission Detection in Machine Translation}, 
      author={David Dale and Elena Voita and Janice Lam and Prangthip Hansanti and Christophe Ropers and Elahe Kalbassi and Cynthia Gao and Loïc Barrault and Marta R. Costa-jussà},
      year={2023},
      eprint={2305.11746},
      archivePrefix={arXiv},
      primaryClass={cs.CL},
      url={https://arxiv.org/abs/2305.11746}, 
}

@article{guerreiro-etal-2023-hallucinations,
    title = "Hallucinations in Large Multilingual Translation Models",
    author = "Guerreiro, Nuno M.  and
      Alves, Duarte M.  and
      Waldendorf, Jonas  and
      Haddow, Barry  and
      Birch, Alexandra  and
      Colombo, Pierre  and
      Martins, Andr{\'e} F. T.",
    journal = "Transactions of the Association for Computational Linguistics",
    volume = "11",
    year = "2023",
    address = "Cambridge, MA",
    publisher = "MIT Press",
    url = "https://aclanthology.org/2023.tacl-1.85/",
    doi = "10.1162/tacl_a_00615",
    pages = "1500--1517"
}

@inproceedings{DBLP:conf/acl/GuerreiroCPM23,
  author       = {Nuno Miguel Guerreiro and
                  Pierre Colombo and
                  Pablo Piantanida and
                  Andr{\'{e}} F. T. Martins},
  editor       = {Anna Rogers and
                  Jordan L. Boyd{-}Graber and
                  Naoaki Okazaki},
  title        = {Optimal Transport for Unsupervised Hallucination Detection in Neural
                  Machine Translation},
  booktitle    = {Proceedings of the 61st Annual Meeting of the Association for Computational
                  Linguistics (Volume 1: Long Papers), {ACL} 2023, Toronto, Canada,
                  July 9-14, 2023},
  pages        = {13766--13784},
  publisher    = {Association for Computational Linguistics},
  year         = {2023},
  url          = {https://doi.org/10.18653/v1/2023.acl-long.770},
  doi          = {10.18653/V1/2023.ACL-LONG.770},
  bibsource    = {dblp computer science bibliography, https://dblp.org}
}

@inproceedings{dou2021word,
  title={Word Alignment by Fine-tuning Embeddings on Parallel Corpora},
  author={Dou, Zi-Yi and Neubig, Graham},
  booktitle={Conference of the European Chapter of the Association for Computational Linguistics (EACL)},
  year={2021}
}

@misc{feng2022languageagnosticbertsentenceembedding,
      title={Language-agnostic BERT Sentence Embedding}, 
      author={Fangxiaoyu Feng and Yinfei Yang and Daniel Cer and Naveen Arivazhagan and Wei Wang},
      year={2022},
      eprint={2007.01852},
      archivePrefix={arXiv},
      primaryClass={cs.CL},
      url={https://arxiv.org/abs/2007.01852}, 
}

@inproceedings{ferrando-etal-2022-towards,
    title = "Towards Opening the Black Box of Neural Machine Translation: Source and Target Interpretations of the Transformer",
    author = "Ferrando, Javier  and
      G{\'a}llego, Gerard I.  and
      Alastruey, Belen  and
      Escolano, Carlos  and
      Costa-juss{\`a}, Marta R.",
    editor = "Goldberg, Yoav  and
      Kozareva, Zornitsa  and
      Zhang, Yue",
    booktitle = "Proceedings of the 2022 Conference on Empirical Methods in Natural Language Processing",
    month = dec,
    year = "2022",
    address = "Abu Dhabi, United Arab Emirates",
    publisher = "Association for Computational Linguistics",
    url = "https://aclanthology.org/2022.emnlp-main.599/",
    doi = "10.18653/v1/2022.emnlp-main.599",
    pages = "8756--8769"
}

@article{Ferrando2022MeasuringTM,
  title={Measuring the Mixing of Contextual Information in the Transformer},
  author={Javier Ferrando and Gerard I. G{\'a}llego and Marta Ruiz Costa-juss{\`a}},
  journal={ArXiv},
  year={2022},
  volume={abs/2203.04212},
  url={https://api.semanticscholar.org/CorpusID:247315171}
}

@article{Sui2024ConfabulationTS,
  title={Confabulation: The Surprising Value of Large Language Model Hallucinations},
  author={Peiqi Sui and Eamon Duede and Sophie Wu and Richard Jean So},
  journal={ArXiv},
  year={2024},
  volume={abs/2406.04175},
  url={https://api.semanticscholar.org/CorpusID:270285964}
}

@misc{hendy2023goodgptmodelsmachine,
      title={How Good Are GPT Models at Machine Translation? A Comprehensive Evaluation}, 
      author={Amr Hendy and Mohamed Abdelrehim and Amr Sharaf and Vikas Raunak and Mohamed Gabr and Hitokazu Matsushita and Young Jin Kim and Mohamed Afify and Hany Hassan Awadalla},
      year={2023},
      eprint={2302.09210},
      archivePrefix={arXiv},
      primaryClass={cs.CL},
      url={https://arxiv.org/abs/2302.09210}, 
}

@misc{raunak2022saltedframeworksalientlongtail,
      title={SALTED: A Framework for SAlient Long-Tail Translation Error Detection}, 
      author={Vikas Raunak and Matt Post and Arul Menezes},
      year={2022},
      eprint={2205.09988},
      archivePrefix={arXiv},
      primaryClass={cs.CL},
      url={https://arxiv.org/abs/2205.09988}, 
}

@inproceedings{yankovskaya-etal-2018-quality,
    title = "Quality Estimation with Force-Decoded Attention and Cross-lingual Embeddings",
    author = {Yankovskaya, Elizaveta  and
      T{\"a}ttar, Andre  and
      Fishel, Mark},
    editor = "Bojar, Ond{\v{r}}ej  and
      Chatterjee, Rajen  and
      Federmann, Christian  and
      Fishel, Mark  and
      Graham, Yvette  and
      Haddow, Barry  and
      Huck, Matthias  and
      Yepes, Antonio Jimeno  and
      Koehn, Philipp  and
      Monz, Christof  and
      Negri, Matteo  and
      N{\'e}v{\'e}ol, Aur{\'e}lie  and
      Neves, Mariana  and
      Post, Matt  and
      Specia, Lucia  and
      Turchi, Marco  and
      Verspoor, Karin",
    booktitle = "Proceedings of the Third Conference on Machine Translation: Shared Task Papers",
    month = oct,
    year = "2018",
    address = "Belgium, Brussels",
    publisher = "Association for Computational Linguistics",
    url = "https://aclanthology.org/W18-6466/",
    doi = "10.18653/v1/W18-6466",
    pages = "816--821"
}

@misc{benkirane2024machinetranslationhallucinationdetection,
      title={Machine Translation Hallucination Detection for Low and High Resource Languages using Large Language Models}, 
      author={Kenza Benkirane and Laura Gongas and Shahar Pelles and Naomi Fuchs and Joshua Darmon and Pontus Stenetorp and David Ifeoluwa Adelani and Eduardo Sánchez},
      year={2024},
      eprint={2407.16470},
      archivePrefix={arXiv},
      primaryClass={cs.CL},
      url={https://arxiv.org/abs/2407.16470}, 
}

@article{Hammadi_Yagi_Fareh_2025, title={Explicitation and Implicitation in Arabic- English Translation of Institutional Academic Correspondence }, volume={25}, url={https://ijaes2011.net/index.php/IJAES/article/view/678}, DOI={10.33806/ijaes.v25i1.678}, number={1}, journal={International Journal of Arabic-English Studies}, author={Al Hammadi, Nada Mohamed and Yagi, Sane Mo and Fareh, Shehdeh}, year={2025}, month={Jan.}, pages={239–258} }

@inproceedings{vazquez-etal-2025-semeval,
    title = "{S}em{E}val-2025 Task 3: Mu-{SHROOM}, the Multilingual Shared-task on Hallucinations and Related Observable Overgeneration Mistakes",
    author = {Vazquez, Raul  and
      Mickus, Timothee  and
      Zosa, Elaine  and
      Vahtola, Teemu  and
      Tiedemann, J{\"o}rg  and
      Sinha, Aman  and
      Segonne, Vincent  and
      Sanchez - Vega, Fernando  and
      Raganato, Alessandro  and
      Libovick{\'y}, Jind{\v{r}}ich  and
      Karlgren, Jussi  and
      Ji, Shaoxiong  and
      Helcl, Jind{\v{r}}ich  and
      Guillou, Liane  and
      De Gibert, Ona  and
      Bengoetxea, Jaione  and
      Attieh, Joseph  and
      Apidianaki, Marianna},
    editor = "Rosenthal, Sara  and
      Ros{\'a}, Aiala  and
      Ghosh, Debanjan  and
      Zampieri, Marcos",
    booktitle = "Proceedings of the 19th International Workshop on Semantic Evaluation (SemEval-2025)",
    month = jul,
    year = "2025",
    address = "Vienna, Austria",
    publisher = "Association for Computational Linguistics",
    url = "https://aclanthology.org/2025.semeval-1.322/",
    pages = "2472--2497",
    ISBN = "979-8-89176-273-2"
}

@inproceedings{mickus-etal-2024-semeval,
    title = "{S}em{E}val-2024 Task 6: {SHROOM}, a Shared-task on Hallucinations and Related Observable Overgeneration Mistakes",
    author = {Mickus, Timothee  and
      Zosa, Elaine  and
      Vazquez, Raul  and
      Vahtola, Teemu  and
      Tiedemann, J{\"o}rg  and
      Segonne, Vincent  and
      Raganato, Alessandro  and
      Apidianaki, Marianna},
    editor = {Ojha, Atul Kr.  and
      Do{\u{g}}ru{\"o}z, A. Seza  and
      Tayyar Madabushi, Harish  and
      Da San Martino, Giovanni  and
      Rosenthal, Sara  and
      Ros{\'a}, Aiala},
    booktitle = "Proceedings of the 18th International Workshop on Semantic Evaluation (SemEval-2024)",
    month = jun,
    year = "2024",
    address = "Mexico City, Mexico",
    publisher = "Association for Computational Linguistics",
    url = "https://aclanthology.org/2024.semeval-1.273/",
    doi = "10.18653/v1/2024.semeval-1.273",
    pages = "1979--1993"
}
\appendix

\section{Missed minimally detached overgenerations}\label{appendix:FNs}
Included below in Table \ref{FNs} are overgenerations from the minimally detached dataset which were not detected due to the fact that the minimum span in this experiment was set to 2. Lowering it to 1 results in too many false positives. The overgenerated ngram is italicized.

\begin{table*}
  \centering
    \begin{tabular}{>{\small}l|>{\small}l}
    \hline 
    \textbf{src} & ``Outside, there's a balcony or terrace offering a view of the sea. ''	\\
    \hline 
    \textbf{tgt} & ``All'esterno, c'è un balcone o una terrazza che offre una vista \emph{mozzafiato} sul mare.''\\
    \hline 
    \hline 
    \textbf{src} & ``The large, color screen makes it easy to view your measurements and calculate area and volume.''	\\
    \hline 
    \textbf{tgt} & ``Lo schermo a colori grande \emph{e luminoso} rende facile la visualizzazione delle misurazioni e il calcolo dell'area e del \\ &volume.''\\
    \hline 
    \hline 
    \textbf{src} & ``At least give us an option to see the newest comments first like YT does.''	\\
    \hline 
    \textbf{tgt} & ``\emph{Per favore}, dateci la possibilità di vedere prima i commenti più recenti come fa YT.''\\
    \hline 
    \hline 
    \textbf{src} & ``With side hand pockets and a soft, smooth waistband—these are it.''	\\
    \hline 
    \textbf{tgt} & ``Con tasche laterali e una cintura liscia e morbida, questi \emph{pantaloni} sono perfetti.''	\\
    \hline
    \end{tabular}
    \caption{\label{FNs}
    Overgenerations from the minimally detached dataset which were not detected due to the fact that the minimum span in this run was set to 2. Lowering it to 1 results in too many false positives. Overgenerated ngram is italicized for clarity.
  }
\end{table*}

\section{Examples of Explicitation}\label{appendix:explicitations}

In Tables \ref{explicitation1} -  \ref{explicitation} we include the full text for the explicitation examples mentioned earlier.
\begin{table*}
  \centering
    \begin{tabular}{>{\small}l|>{\small}l}
    \hline 
    \textbf{src} & ``The Department of Justice's overall approach to corporate crimes has come under intense scrutiny from\\& advocates. While watchdogs have lauded some steps the department has taken during Merrick Garland's\\& tenure
    as attorney general - like finally establishing a database on corporate crime - Biden's \emph{DOJ} has \\& also leaned
    heavily on leniency agreements that allow companies to defer or avoid prosecution, and\\& encouraged companies to 
    scapegoat individual employees so as to avoid a broader charge. Kenneth Polite\\& Jr., former assistant
    attorney general for DOJ's Criminal Division, revised the division's corporate \\&enforcement policy so as to limit prosecutions.''	\\
    \hline 
    \textbf{tgt} & ``L'approccio generale del Dipartimento di Giustizia ai reati societari è stato sottoposto a un intenso scrutinio \\&da parte degli avvocati. Mentre i cani da guardia hanno elogiato alcuni passi intrapresi dal Dipartimento \\&durante il mandato di Merrick Garland come procuratore generale, come l'istituzione di un database sui reati \\& societari,
    \emph{il Dipartimento} \emph{di Giustizia} di Biden si è anche affidato pesantemente agli accordi di clemenza che\\& consentono alle società di differire o evitare l'accusa, 
    e ha incoraggiato le società a incolpare singoli \\& dipendenti  al fine di evitare un'accusa più ampia. Kenneth Polite Jr., ex assistente procuratore generale della\\& Divisione penale del Dipartimento di Giustizia, ha rivisto la politica di applicazione delle norme societarie \\&della divisione in modo da limitare le azioni penali.'' \\
    \hline 
    detected OG & [``di Giustizia di'']\\
    \end{tabular}
    \caption{\label{explicitation1}
    Example of explicitation. Target text has expanded the ``DOJ'' acronym, where the English source used it after first mention. (italics added)
  }
\end{table*}

\begin{table*}
  \centering
    \begin{tabular}{>{\small}l|>{\small}l}
    \hline 
    \textbf{src} & An 83-hour wait in a hospital \emph{A\&E}; four-in-ten patients waiting longer than four hours; 30 ambulances \\& queuing outside a hospital, unable to respond to other emergency calls; staff ``leaving shifts in tears'' over the \\& standard of care and ``a near collapse of performance,'' according to an expert at the Royal College of\\& Emergency Medicine.\\
    \hline 
    \textbf{tgt} &Un'attesa di 83 ore \emph{al pronto soccorso} di un ospedale; quattro pazienti su dieci aspettano più di quattro ore; \\& 30 ambulanze in coda fuori da un ospedale, impossibilitate a rispondere ad altre chiamate di emergenza; \\& il personale ``lascia il turno in lacrime'' a causa della qualità delle cure e ``quasi un collasso delle prestazioni'',\\& secondo un esperto del Royal College of Emergency Medicine.	\\
     \hline 
    detected OG &  [``al pronto soccorso'']\\
 \end{tabular}
    \caption{\label{explicitation}
    Example of explicitation. Target text has expanded ``hospital A\&E'' from source to ``al pronto soccorso di un ospedale''.
  }
\end{table*}

\begin{table*}
  \centering
    \begin{tabular}{>{\small}l|>{\small}l}
    \hline 
    \textbf{src} & Our planning system also needs to adapt so that planners in suburban councils, who seem to be less supportive\\& of change, are not the only determinants of the future shape of their suburbs. A way forward could be to have\\& an independent decision-making body that championed quality design as long as it fitted below the tree tops\\& and incorporated family living, large balconies, common gardens and presented a \\& friendly appearance. This could be a role for the current NSW government architect working within a \\&carefully crafted complying code.\\
    \hline
    \textbf{tgt} & Anche il nostro sistema di pianificazione deve adattarsi in modo che i pianificatori dei consigli dei sobborghi,\\& che sembrano essere meno favorevoli al cambiamento, non siano gli unici a determinare la forma futura dei \\&loro sobborghi. Un modo per procedere potrebbe essere quello di avere un organismo\\& decisionale indipendente che promuova un design di qualità, purché si adatti sotto le cime degli alberi e \\&integri la vita familiare, ampi balconi, giardini comuni e presenti un aspetto amichevole. Questo potrebbe\\& essere un ruolo per l'attuale architetto del governo del New South Wales che lavora all'interno di \\&un codice di conformità attentamente elaborato.	\\
    \hline
    detected OG &    [``del New South'']\\

\end{tabular}
    \caption{\label{explicitation}
    Example of explicitation. Target text has expanded ``NSW'' from source to ``del New South Wales''.
  }
\end{table*}  

\end{document}